%% file: main.tex
\documentclass[nohyperref]{article}

\input{math_commands.tex}

\usepackage{hyperref}
\usepackage{url}
\usepackage{xcolor}
\usepackage{threeparttable}

\usepackage{microtype}
\usepackage{graphicx}
\usepackage{subfigure}
\usepackage{booktabs} %

\usepackage{amsmath, amssymb, amsthm}
\usepackage{graphicx}
\usepackage{booktabs}
\usepackage{subfigure}
\usepackage{caption}
\usepackage{enumitem}
\setitemize{noitemsep,topsep=0pt,parsep=0pt,partopsep=0pt}
\usepackage{multicol, multirow}
\usepackage{tikz, pgfplots}
\usetikzlibrary{matrix}
\usepgfplotslibrary{groupplots}
\pgfplotsset{compat=newest}

\definecolor{red}{HTML}{E31A1C}
\definecolor{blue}{HTML}{1F78B4}
\definecolor{green}{HTML}{33A02C}
\definecolor{orange}{HTML}{FF7F00}
\definecolor{purple}{HTML}{6A3D9A}

\usepackage{times}
\usepackage{natbib}

\newcommand{\xhdr}[1]{{\noindent\bfseries #1}.}

\newcommand{\bfg}[1]{\textbf{\color{green}#1}}
\newcommand{\bfr}[1]{\textbf{\color{red}#1}}

\newcommand{\fone}{F\textsubscript{1}}

\newcommand{\methodname}{\textsc{SpanDrop}}
\newcommand{\spanmask}{\textsc{SpanMask}}
\newcommand{\toytask}{\textsc{FindAnimals}}
\newcommand{\toytasksmall}{\textsc{FindCats}}

\newcommand{\eg}{\textit{e.g.}}
\newcommand{\ie}{\textit{i.e.}}

\newtheorem{definition}{Definition}

\newtheorem{remark}{Remark}

\usepackage[belowskip=-12pt,aboveskip=7pt]{caption}

\newcommand{\B}{\mathrm{B}}

\begin{document}
\twocolumn

\title{\methodname{}: Simple and Effective Counterfactual Learning for Long Sequences}

\author{Peng Qi$^{*\dagger1}$\quad Guangtao Wang$^{*\dagger2}$\quad Jing Huang$^{\dagger3}$\\
{\small $^1$ AWS AI, Seattle, WA \quad$^2$ TikTok, Mountain View, CA\quad $^3$ Alexa AI, Sunnyvale, CA}\\
{\tt\small pengqi@cs.stanford.edu}
}

\date{}
\maketitle

\renewcommand*{\thefootnote}{\fnsymbol{footnote}}
\setcounter{footnote}{1}
\footnotetext{Equal contribution.}
\setcounter{footnote}{2}
\footnotetext{Work done at JD AI Research.}
\renewcommand*{\thefootnote}{\arabic{footnote}}
\setcounter{footnote}{0}

\begin{abstract}
Distilling supervision signal from a long sequence to make predictions is a challenging task in machine learning, especially when not all elements in the input sequence contribute equally to the desired output.
In this paper, we propose \methodname{}, a simple and effective data augmentation technique that helps models identify the true supervision signal in a long sequence with very few examples.
By directly manipulating the input sequence, \methodname{} randomly ablates parts of the sequence at a time and ask the model to perform the same task to emulate counterfactual learning and achieve input attribution.
Based on theoretical analysis of its properties, we also propose a variant of \methodname{} based on the beta-Bernoulli distribution, which yields diverse augmented sequences while providing a learning objective that is more consistent with the original dataset.
We demonstrate the effectiveness of \methodname{} on a set of carefully designed toy tasks, as well as various natural language processing tasks that require reasoning over long sequences to arrive at the correct answer, and show that it helps models improve performance both when data is scarce and abundant.
\end{abstract}

\section{Introduction}

Building effective machine learning systems for long sequences is a challenging and important task, which helps us better understand underlying patterns in naturally occurring sequential data like long texts \citep{radford2019language}, protein sequences \citep{jumper2021highly}, financial time series \citep{bao2017deep}, etc.
Recently, %
there is growing interest in studying neural network models that can capture long-range correlations in sequential data with high computational, memory, and statistical efficiency, especially widely adopted Transformer models
\citep{vaswani2017attention}. %

Previous work approach long-sequence learning in Transformers largely by introducing computational approaches to replace the attention mechanism with more efficient counterparts.
These approaches include limiting the scope of the attention mechanism \citep{kitaev2020reformer},  limiting sequence-level attention to only a handful of positions \citep{beltagy2020longformer, zaheer2020big}, or borrowing ideas from the kernel trick to eliminate the need to compute or instantiate the costly attention matrix \citep{peng2021random, katharopoulos2020transformers, choromanski2020rethinking}.
Essentially, these approaches aim to approximate the original pair-wise interaction with lower cost, and are often interested in capturing the effect of every input element on the outcome \citep[\eg, arithmetic operations over a long list of numbers and operators, as proposed by][]{tay2020long}.
While these tasks present great challenges for designing effective models to handle long sequences, many real-world problems involving long sequences share properties that make them amenable to more effective approaches than modeling the raw input-output relationship directly (\eg, via decomposition).

\begin{figure}
    \centering
    \includegraphics[width=0.65\linewidth]{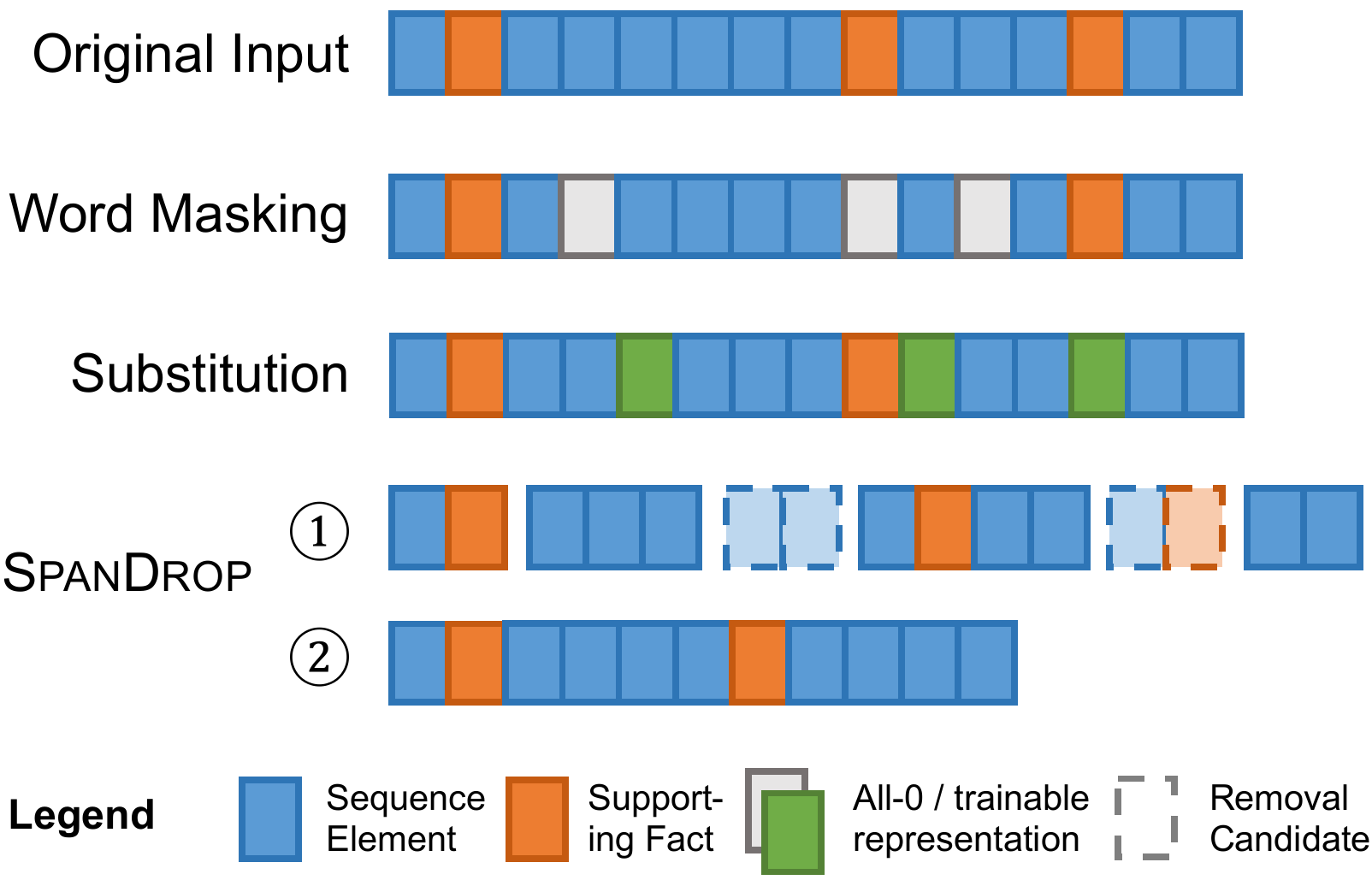}
    \caption{\methodname{} compared to masking- and substitution-based augmentation techniques (\eg, word dropout) commonly used in sequence learning. 
    \methodname{} can not only work with arbitrary granularities of input units, but also introduce augmented sequences without artificial tokens and alleviates potential positional bias in the data.}
    \label{fig:spandrop_vs_others}
\end{figure}

In this paper, we focus on a learning problem for long sequences motivated by real-world tasks, where not all input elements might contribute to the desired output. 
Natural examples that take this form include sentiment classification for long customer review documents (where a few salient sentiment words and conjunctions contribute the most), question answering from a long document (where each question typically requires a small number of supporting sentences to answer), key phrase detection in audio processing (where a small number of recorded frames actually determine the prediction), as well as detecting a specific object from a complex scene (where, similarly, a small amount of pixels determine the outcome), to name a few.
In these problems, it is usually counterproductive to try and make direct use of the entire input if the contributing portion is small or sparse, which results in a problem of \emph{underspecification} (i.e., the training data does not sufficiently define the goal for statistical models).

One approach to address this problem is to annotate the parts of input that directly contribute to the outcome. This could take the form of a subset of sentences that answer a question or describe the relation between entities in a paragraph \citep{yang2018hotpotqa,yao2019docred}.
However, such annotation is not always technically or financially feasible, so researchers and practitioners often need to resort to either collecting more input-output pairs or designing problem-specific data augmentation techniques to make up for the data gap.
For real-valued data, augmentation often translates to random transformations (\eg, shifting or flipping an image);
for symbolic data like natural language, techniques like masking or substitution are more commonly used (\eg, randomly swapping words with a special mask token or other words). 
While these approaches have proven effective in some tasks, each has limitations that prevents it from being well-suited for the underspecification scenario.
For instance, while global feature transformations enhance group-invariance in learned representations, they do not directly help with better locating the underlying true stimulus.
On the other hand, while replacement techniques like masking and substitution help ablate parts of the input, they are susceptible to the position bias of where the true stimulus might occur in the input.
Furthermore, while substitution techniques can help create challenging contrastive examples, they are often significantly more difficult to implement (\eg, replacing a phrase in a sentence without losing fluency).

To address these challenges, we propose \methodname{}, a simple and effective technique that helps models distill sparse supervision signal from long sequences when the problem is underspecified.
\methodname{} randomly ablates parts of the input to construct \emph{counterfactual} examples that preserves the original output and supervision signal with high probability.
Unlike replacement-based techniques, however, \methodname{}  removes ablated elements from the input and concatenate the remainder.
This avoids introducing artificial representations that are not used at test time, and mitigates potential spurious correlation to absolute positions of outcome-determining parts of the input (see Figure \ref{fig:spandrop_vs_others}).
We further propose a theoretically motivated, more effective variant of \methodname{} based on the beta-Bernoulli distribution that enhances the consistency of the augmented objective function with the original one.
We demonstrate via carefully designed toy experiments that \methodname{} not only helps models achieve up to $20\times$ sample-efficiency in low-data settings, but also further reduces overfitting even when training data is abundant.
We find that it is very effective at mitigating position bias compared to replacement-based counterfactual approaches, and enhances out-of-distribution generalization effectively.
We further experiment on four natural language processing tasks that require models to answer question or extract entity relations from long texts, where, motivated by our theoretical analysis, we further propose an adaptive span segmentation. 
Our experiments demonstrate that \methodname{} can improve the performance of competitive neural models without any architectural change.

To summarize, our contributions in this paper are: 1) we propose \methodname{} and Beta-\methodname{}, two effective approaches to generate counterfactual examples for tasks of long sequence learning; 2) we study the theoretical properties of these approaches and verify them with carefully designed experiments on synthetic data; 3) we demonstrate that both approaches improve upon strong neural models on four NLP datasets, which can be furthered by a theoretically motivated span segmentation approach we propose.

\section{Method}
\input{method}

\section{\toytask{}: Distilling Supervision from Long-Sequences}
\input{toytask.tex}

\section{Experiments on Natural Language Data}
\input{experiments}

\section{Related Work}

\input{relatedwork}

\section{Conclusion}

In this paper, we presented \methodname{}, a simple and effective method for learning from long sequences, which ablates parts of the sequence at random to generate counterfactual data to distill the sparse supervision signal that is predictive of the desired output.
We show via theoretical analysis and carefully designed synthetic datasets that \methodname{} and its variant based on the beta-Bernoulli distribution, Beta-\methodname{}, help models achieve competitive performance with a fraction of the data by introducing diverse augmented training examples, and generalize better to previously unseen data.
Our experiments on four real-world NLP datasets confirm these theoretical findings, and demonstrate \methodname{}'s efficacy on strong neural models even when data is abundant.

\bibliography{mybib}
\bibliographystyle{iclr2022_conference}

\clearpage
\newpage
\appendix
\section{Appendix}
\input{appendix}

\end{document}

%% file: math_commands.tex
\usepackage{amsmath,amsfonts,bm}

\def\eqref#1{equation~\ref{#1}}

\def\1{\bm{1}}

\DeclareMathAlphabet{\mathsfit}{\encodingdefault}{\sfdefault}{m}{sl}
\SetMathAlphabet{\mathsfit}{bold}{\encodingdefault}{\sfdefault}{bx}{n}

\newcommand{\E}{\mathbb{E}}

%% file: method.tex
In this section, we first formulate the problem of sequence inference, where the model takes sequential data as input to make predictions. Then, we introduce \methodname{}, a simple and effective data augmentation technique for long sequence inference, and analyze its theoretical properties.

\subsection{Problem Definition}

\xhdr{Sequence Inference} We consider a task where a model takes a sequence $\bm{S}$ as input and predicts the output $\bm{y}$.
We assume that $\bm{S}=(\bm{s}_1, \ldots, \bm{s}_n)$ consists of $n$ disjoint but contiguous \emph{spans}, and each span represents a part of the sequence in order.
One example of sequence inference is sentiment classification from a paragraph of text, where $\bm{S}$ is the paragraph and $\bm{y}$ the sentiment label.
Spans could be words, phrases, sentences, or a mixture of these in the paragraph.
Another example is time series prediction, where $\bm{S}$ is historical data, $\bm{y}$ is the value at the next time step.

\xhdr{Supporting Facts} Given an input-output pair $(\bm{S}, \bm{y})$ for sequence prediction, we assume that $\bm{y}$ is truly determined by only a subset of spans in $\bm{S}$.
More formally, we assume that there is a subset of spans $S_\mathrm{sup} \subset \{\bm{s}_1, \bm{s}_2, \ldots, \bm{s}_n\}$ such that $\bm{y}$ is independent of $\bm{s}_i$, if $\bm{s_i} \notin S_\mathrm{sup}$.
In sentiment classification, $S_\mathrm{sup}$ could consist of important sentiment words or conjunctions (like ``good'', ``bad'', ``but''); in time series prediction, it could reflect the most recent time steps as well as those a few cycles away if the series is periodic. 
For simplicity, we will denote the size of this set $m=|S_\mathrm{sup}|$, and restrict our attention to tasks where $m \ll n$, such as those described in the previous section.

\subsection{\methodname{}}

In a long sequence inference task with sparse support facts ($m \ll n$), most of the spans in the input sequence will not contribute to the prediction of $\bm{y}$, but they will introduce spurious correlation in a low-data scenario.
\methodname{} generates new data instances $(\tilde{\bm{S}}, \bm{y})$ by ablating these spans at random, while preserving the supporting facts with high probability so that the model is still trained to make the correct prediction $\bm{y}$.
This is akin to counterfactually determining whether each span truly determines the outcome $\bm{y}$ by asking what the prediction would have been without it.

\begin{definition}[\methodname{}]
Formally, given a sequence $\bm{S}$ that consists of spans $(\bm{s}_1, \bm{s}_2, \cdots \bm{s}_n)$, \methodname{} generates a new sequence $\tilde{\bm{S}}$ as follows:
\vspace{-.5em}
\begin{align*}
    \delta_i \overset{i.i.d.}{\sim}& \mathrm{Bernoulli}(1-p),& \tilde{\bm{S}} = & (\bm{s}_i)_{i=1, \delta_i =1 }^{n},
\end{align*}
where $p$ is the hyperparameter that determines the probability to drop a span.
\end{definition}

Note that \methodname{} does not require introducing substitute spans or artificial symbols when ablating spans from the input sequence.
It makes the most of the natural sequence as it occurs in the original training data, and preserves the relative order between spans that are not dropped, which is often helpful in understanding sequential data (e.g., time series or text).
It is also not difficult to establish that the resulting sequence $\tilde{\bm{S}}$ can preserve all of the $m$ supporting facts with high probability regardless of how large $n$ is.
\label{remark:bernoulli}
\begin{remark}
The new sequence length $n'=|\tilde{\bm{S}}|$ and the number of preserved supporting facts $m'=|\tilde{\bm{S}}\cap S_\mathrm{sup}|$ follow binomial distributions $\mathrm{Bin}(n,p)$ and $\mathrm{Bin}(m, p)$, respectively, where $P(x=k|N,p)={N \choose k}(1-p)^{k}p^{N-k}$ for $X\sim \mathrm{Bin}(N, p)$.
\end{remark}
Therefore, the proportion of sequences where all supporting facts are retained (\ie, $m'=m$) is $(1-p)^m$, which is independent of $n$.
This means that as long as the total number of supporting facts in the sequence is bounded, then regardless of the sequence length, we can always choose $p$ carefully such that we end up with many valid new examples with bounded noise introduced to supporting facts.
Note that our analysis so far relies only on the assumption that $m$ is known or can be estimated, and thus it can be applied to tasks where the precise set of supporting facts $S_\mathrm{sup}$ is unknown.
More formally, the amount of new examples can be characterized by the size of the \emph{typical set} of $\tilde{\bm{S}}$, \ie,  the set of sequences that the randomly ablated sequence will fall into with high probability.
The size of the typical set for \methodname{} is approximately $2^{nH(p)}$, where $H(p)$ is the binary entropy of a Bernoulli random variable with probability $p$.
Intuitively, these results indicate that the amount of total counterfactual examples generated by \methodname{} scales exponentially in $n$, but the level of supporting fact noise can be bounded as long as $m$ is small.

\begin{figure}
    \centering
    \pgfplotstableread{
        l   d   b
        0	0.0	0.0
        1	0.0	0.0
        2	0.0	0.0
        3	0.0	0.0006000000000000001
        4	0.0	0.0007
        5	0.0	0.0012000000000000001
        6	0.0	0.0022
        7	0.0	0.0019000000000000002
        8	0.0	0.004
        9	0.0	0.0046
        10	0.0	0.0077
        11	0.0	0.007600000000000001
        12	0.0	0.0081
        13	0.0	0.0119
        14	0.0	0.015799999999999998
        15	0.0	0.0177
        16	0.0	0.020900000000000002
        17	0.0	0.0238
        18	0.0	0.0297
        19	0.0	0.0299
        20	0.0	0.0344
        21	0.0	0.0441
        22	0.0	0.049
        23	0.0	0.0582
        24	0.0	0.0649
        25	0.0	0.0703
        26	0.0	0.0816
        27	0.0	0.08750000000000001
        28	0.0	0.0961
        29	0.0	0.1117
        30	0.0	0.12340000000000001
        31	0.0	0.13040000000000002
        32	0.0	0.14200000000000002
        33	0.0	0.1569
        34	0.0	0.168
        35	0.0	0.1759
        36	0.0	0.1926
        37	0.0	0.21619999999999998
        38	0.0	0.2345
        39	0.0	0.2424
        40	0.0	0.2748
        41	0.0	0.2944
        42	0.0	0.3097
        43	0.0	0.3291
        44	0.0	0.36110000000000003
        45	0.0	0.37429999999999997
        46	0.0	0.3972
        47	0.0	0.4339
        48	0.0	0.4601
        49	0.0	0.48849999999999993
        50	0.0	0.5124
        51	0.0	0.5414
        52	0.0	0.5721999999999999
        53	0.0	0.6019
        54	0.0	0.6476
        55	0.0	0.6764
        56	0.0	0.7041000000000001
        57	0.0	0.7378
        58	9.999999999999999e-05	0.7691
        59	0.00019999999999999998	0.8258
        60	0.00019999999999999998	0.853
        61	0.001	0.902
        62	0.0018	0.9611
        63	0.0025	0.9942
        64	0.0104	1.0474
        65	0.016	1.0826
        66	0.036000000000000004	1.1373
        67	0.08009999999999999	1.1927999999999999
        68	0.15430000000000002	1.2501
        69	0.2964	1.294
        70	0.5192	1.3649
        71	0.8702	1.4118
        72	1.4112	1.4384
        73	2.1784999999999997	1.5114
        74	3.1779	1.6004
        75	4.3453	1.6558
        76	5.7759	1.7287000000000001
        77	7.2257	1.7781000000000002
        78	8.488199999999999	1.8789
        79	9.4774	1.9366
        80	9.9273	2.0044
        81	9.786	2.077
        82	9.0658	2.1414
        83	7.8721	2.2422
        84	6.3768	2.3022
        85	4.8383	2.3869000000000002
        86	3.3606999999999996	2.4737
        87	2.1642	2.543
        88	1.2839	2.6412999999999998
        89	0.6848	2.7196000000000002
        90	0.3368	2.8277
        91	0.1471	2.9192
        92	0.0588	3.0305
        93	0.0223	3.1038
        94	0.0045000000000000005	3.1863
        95	0.0015	3.3013
        96	0.00039999999999999996	3.3821999999999997
        97	9.999999999999999e-05	3.524
        98	0.0	3.6109
        99	9.999999999999999e-05	3.7409999999999997
        100	0.0	3.8417
    }\lengthdistn
    
    \pgfplotstableread[row sep=crcr]{
    l mtwo mone zero one two max \\
    1	0.32508297	0.32508297	0.32508297	0.32508297	0.32508297 0.32508297\\
10	0.07259595	0.20660956	0.31403354	0.32487793	0.32508075 0.32508297\\
100	0.03309127	0.1571845	0.29405124	0.32360209	0.32505983 0.32508297\\
1000	0.02754634	0.1466072	0.28510245	0.32136036	0.32493009 0.32508297\\
10000	0.02680593	0.14487911	0.28323173	0.32039728	0.32470366 0.32508297\\
100000	0.02671071	0.14463161	0.28294182	0.32020163	0.32460689 0.32508297\\
    }\spanentropy
    
    \pgfplotstableread[row sep=crcr]{
    l mtwo mone zero one two max \\
    1 -0.22314355131420927 -0.22314355131420982 -0.22314355131420927 -0.22314355131420882 -0.22314355131425145 -0.2231435477733612 \\
2 -0.2327130023303603 -0.2921364228011615 -0.40546510810816416 -0.4413971173342617 -0.4457882250919738 -0.4462870657444 \\
3 -0.2376029876245518 -0.33295841732141673 -0.5596157879354231 -0.6549712176323368 -0.6679362561651487 -0.6694304943084717 \\
4 -0.2408870628257409 -0.36194595419466846 -0.6931471805599458 -0.8640630154908564 -0.8895898643179407 -0.8925739079713821 \\
5 -0.24335925197113095 -0.38441881004672707 -0.8109302162163279 -1.0688574281368943 -1.1107512544283509 -1.1157172620296478 \\
6 -0.24534141317512237 -0.40276794871492194 -0.9162907318741564 -1.2695281235990166 -1.3314226165969103 -1.3388605564832687 \\
7 -0.24699567327114913 -0.418272135250889 -1.0116009116784785 -1.466238417845105 -1.5516061262831045 -1.5620038211345673 \\
8 -0.24841511992537413 -0.43169515558302807 -1.0986122886681078 -1.6591420839695417 -1.771303944431338 -1.785147026181221 \\
9 -0.24965812816497213 -0.44352961323003137 -1.178654996341646 -1.8483840836080816 -1.9905182176050857 -2.008290156722069 \\
10 -0.2507637114727217 -0.4541117225605683 -1.252762968495372 -2.0341012294031913 -2.2092510781133115 -2.231433257460594 \\
11 -0.25175923171422276 -0.4636811735767198 -1.3217558399823202 -2.216422786197157 -2.4275046441337054 -2.454576328396797 \\
12 -0.25266461882903357 -0.4724148535454761 -1.3862943611198912 -2.3954710176461305 -2.6452810198409225 -2.677719309926033 \\
13 -0.25349483888502533 -0.48044702524273164 -1.446918982936325 -2.571361684109789 -2.8625822955302738 -2.9008622765541077 \\
14 -0.25426141615522013 -0.4878820037302555 -1.504077396776268 -2.744204496949166 -3.0794105477400535 -3.1240051686763763 \\
15 -0.2549734133373107 -0.49480244657482275 -1.5581446180465481 -2.91410353374458 -3.295767839370683 -3.3471480309963226 \\
16 -0.255638086010503 -0.5012749610804471 -1.6094379124340965 -3.0811576184077154 -3.511656219803399 -3.570290818810463 \\
17 -0.2562613331481636 -0.5073540071568482 -1.6582280766035287 -3.2454606696990282 -3.727077725024401 -3.793433576822281 \\
18 -0.2568480154765176 -0.5130846818658193 -1.7047480922384257 -3.407102021255483 -3.94203437773308 -4.016576290130615 \\
19 -0.2574021856209292 -0.518504749335157 -1.7491998548092558 -3.5661667158851316 -4.1565281874688935 -4.239718914031982 \\
20 -0.2579272578304277 -0.5236461488355738 -1.7917594692280479 -3.7227357765767266 -4.370561150711865 -4.46286153793335 \\
    }\supportnoise

    \subfigure[\footnotesize Length of $\tilde{\bm{S}}$ ($n=100, p=0.2$)]{\begin{tikzpicture}[font=\footnotesize,every axis/.append style={scale only axis, axis on top}]
        \begin{axis}[ybar,xlabel={Sequence length ($n'$)},ylabel={Proportion (\%)}, width=6cm, height=1.5cm, bar width=0.005cm,legend cell align=left,legend pos=north west,legend style={nodes={scale=0.67, transform shape}},xtick align=inside]
            \addplot[color=red,fill=red] plot table[x=l, y=d]{\lengthdistn}; 
            \addplot[color=blue,fill=blue] plot table[x=l, y=b]{\lengthdistn}; 
            \legend{\methodname{}, Beta-\methodname{}}
        \end{axis}
    \end{tikzpicture}\label{fig:length_distribution}}\vspace{-1em}
    \\ \subfigure[{\scriptsize Supporting fact noise ($p=0.2$)}]{\begin{tikzpicture}[font=\footnotesize,every axis/.append style={scale only axis, axis on top}]
        \begin{axis}[xlabel={Supporting facts ($m$)},ylabel={Log noise-free prob.}, width=2.75cm, height=2cm, bar width=0.005cm,legend cell align=left,legend style={legend columns=2,nodes={scale=0.45, transform shape}},ymax=3.8]
            \addplot[color=red] plot table[x=l, y=max]{\supportnoise}; 
            \addplot[color=red!70!blue] plot table[x=l, y=two]{\supportnoise}; 
            \addplot[color=red!30!blue] plot table[x=l, y=one]{\supportnoise}; 
            \addplot[color=blue] plot table[x=l, y=zero]{\supportnoise}; 
            \addplot[color=blue!70] plot table[x=l, y=mone]{\supportnoise}; 
            \addplot[color=blue!30] plot table[x=l, y=mtwo]{\supportnoise}; 
            \legend{$\gamma=\infty$, $\gamma=100$, $\gamma=10$, $\gamma=1$, $\gamma=0.1$, $\gamma=0.01$}
        \end{axis}
    \end{tikzpicture}\label{fig:supfact_noise}}
    \ \subfigure[\scriptsize Typical set size ($p=0.1$)]{\begin{tikzpicture}[font=\footnotesize,every axis/.append style={scale only axis, axis on top}]
        \begin{axis}[xlabel={Sequence length ($n$)},ylabel={Avg. nats/span}, width=2.75cm, height=2cm, bar width=0.005cm,legend cell align=left,legend style={legend columns=2,nodes={scale=0.45, transform shape}},xmode=log,ymax=0.6]
            \addplot[color=red] plot table[x=l, y=max]{\spanentropy}; 
            \addplot[color=red!70!blue] plot table[x=l, y=two]{\spanentropy}; 
            \addplot[color=red!30!blue] plot table[x=l, y=one]{\spanentropy}; 
            \addplot[color=blue] plot table[x=l, y=zero]{\spanentropy}; 
            \addplot[color=blue!70] plot table[x=l, y=mone]{\spanentropy}; 
            \addplot[color=blue!30] plot table[x=l, y=mtwo]{\spanentropy}; 
            \legend{$\gamma=\infty$, $\gamma=100$, $\gamma=10$, $\gamma=1$, $\gamma=0.1$, $\gamma=0.01$}
        \end{axis}
    \end{tikzpicture}\label{fig:typical_set}}
    \caption{Theoretical comparison between \methodname{} and Beta-\methodname{}.}
    \label{fig:theoretical_contrast}
\end{figure}
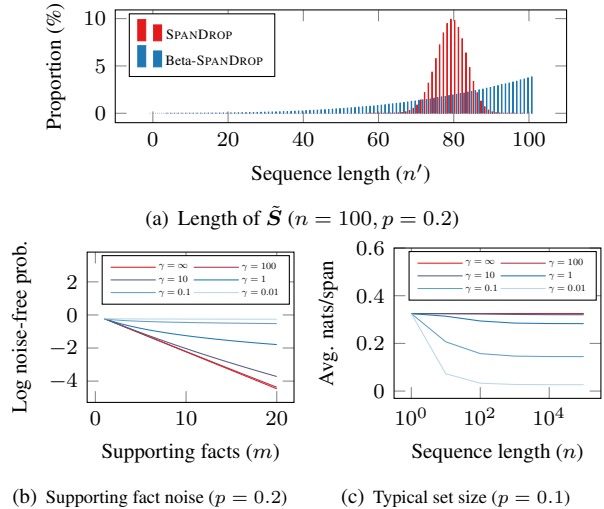

However, this formulation of \methodname{} does have a notable drawback that could potentially hinder its efficacy.
The new sequence length $n'$ follows a binomial distribution, thus for sufficiently large $n$, most $\tilde{\bm{S}}$ lengths will concentrate around the mean $n(1-p)$ with a width of $O(\sqrt{n})$.
This creates an artificial and permanent distribution drift from the original length (see Figure \ref{fig:length_distribution}).
Furthermore, even if we know the identity of $S_\mathrm{sup}$ and keep these spans during training, this length reduction will bias the training set towards easier examples to locate spans in $S_\mathrm{sup}$, potentially hurting generalization performance.
In the next subsection, we will introduce a variant of \methodname{} based on the beta-Bernoulli distribution that alleviates this issue.

\xhdr{Relation to word dropout} A commonly used data augmentation/regularization technique in NLP, word dropout \cite{dai2015semi,gal2016theoretically}, is closely related to \methodname{}.
Two crucial difference, however, set \methodname{} apart from it and similar techniques: 
1) word dropout masks or replaces symbols in the input, while \methodname{} directly removes them. Thus \methodname{} can avoid introducing artificial representations not used at test time, and affect sequence length and the absolute position of remaining elements in the sequence, which we will demonstrate alleviates the effect of spurious correlations between the outcome and absolute element positions in the training data;
2) \methodname{} can operate on input spans segmented at arbitrary granularity (not just words, or even necessarily contiguous), and our theoretical analysis holds as long as these spans are \emph{disjoint}. 
We will show that task-informed span segmentation leads to further performance gains.

\subsection{Beta-\methodname{}}

To address the problem of distribution drift with \methodname{}, we introduce a variant that is based on the beta-Bernoulli distribution.
The main idea is that instead of dropping each span in a sequence independently with a fixed probability $p$, we first sample a sequence-level probability $\pi$ at which spans are dropped from a Beta distribution, then use this probability to perform \methodname{}.

\begin{definition}[Beta-\methodname{}] Let $\alpha = \gamma, \beta = \gamma\cdot\frac{1-p}{p}$, where $\gamma > 0$ is a scaling hyperparameter. Beta-\methodname{} generates $\tilde{\bm{S}}$ over $\bm{S}$ as: 
\vspace{-.5em}
\begin{align*}
    \pi \sim & \B(\alpha, \beta), &
    \delta_i \overset{i.i.d.}{\sim}& \mathrm{Bernoulli}(1-\pi), &
    \tilde{\bm{S}} = & (\bm{s}_i)_{i=1, \delta_i =1}^{n},
\end{align*}
where $\B(\alpha, \beta)$ is the beta-distribution with parameters $\alpha > 0$ and $\beta > 0$.
\end{definition}
It can be easily demonstrated that in Beta-\methodname{}, the probability that each span is dropped is still $p$ on average: $\E[\delta_i|p] = \E[\E[\delta_i|\pi]|p] = \E[1-\pi|p] = 1-\frac{\alpha}{\alpha+\beta}=1-p$.
In fact, we can show that as $\gamma \to \infty$, Beta-\methodname{} degenerates into \methodname{} since the beta-distribution would assign all probability mass on $\pi=p$.
Despite the simplicity in its implementation, Beta-\methodname{} is significantly less likely to introduce unwanted data distribution drift, while is capable of generating diverse counterfactual examples to regularize the training of sequence inference models.
This is due to the following properties:
\begin{remark}
The new sequence length $n'=|\tilde{\bm{S}}|$ and the number of preserved supporting facts $m'=|\tilde{\bm{S}}\cap S_\mathrm{sup}|$ follow beta-binomial distributions $\mathrm{B\mbox{-}Bin}(n, \beta, \alpha)$ and $\mathrm{B\mbox{-}Bin}(m, \beta, \alpha)$, respectively, where \\{\small $P(x=k|N, \alpha, \beta)=\frac{\Gamma(N+1)}{\Gamma(k+1)\Gamma(N-k+1)}\frac{\Gamma(k+\beta)\Gamma(N-k+\alpha)}{\Gamma(N+\alpha+\beta)}\frac{\Gamma(\alpha+\beta)}{\Gamma(\alpha)\Gamma(\beta)}$} for $X\sim \mathrm{B\mbox{-}Bin}(N, \beta, \alpha)$,
and $\Gamma(z)=\int_0^\infty x^{z-1}e^{-x}dx$ the gamma function.
\label{remark:beta}
\end{remark}
As a result, we can show that the probability that Beta-\methodname{} preserves the entire original sequence with the following probability
\begin{align*}
P(n'=n|n,\alpha,\beta)=&\frac{\Gamma(n+\beta)\Gamma(\alpha+\beta)}{\Gamma(n+\alpha+\beta)\Gamma(\beta)}.
\end{align*}
When $\gamma=1$, this expression simply reduces to $\frac{\beta}{n+\beta}$; when $\gamma \ne 1$, this quantity tends to $O(n^{-\gamma})$ as $n$ grows sufficiently large. %
Comparing this to the $O((1-p)^n)$ rate from \methodname{}, we can see that when $n$ is large, Beta-\methodname{} recovers more of the original distribution represented by $(\tilde{\bm{S}}, \bm{y})$ compared to \methodname{}.
In fact, as evidenced by Figure \ref{fig:length_distribution}, the counterfactual sequences generated by Beta-\methodname{} are also more spread-out in their length distribution besides covering the original length $n$ with significantly higher probability.
A similar analysis can be performed by substituting $n$ and $n'$ with $m$ and $m'$, where we can conclude that as $m$ grows, Beta-\methodname{} is much better at generating counterfactual sequences that preserve the entire supporting fact set $S_\mathrm{sup}$.
This is shown in Figure \ref{fig:supfact_noise}, where the proportion of ``noise-free'' examples (\ie, $m'=m$) decays exponentially with \methodname{} ($\gamma=\infty$) while remaining much higher when $\gamma$ is sufficiently small. 
For instance, when $p=0.1$, $\gamma=1$ and $m=10$, the proportion of noise-free examples for \methodname{} is just 34.9\%, while that for Beta-\methodname{} is 47.4\%.

As we have seen, Beta-\methodname{} is significantly better than its Bernoulli counterpart at assigning probability mass to the original data as well as generated sequences that contain the entire set of supporting facts.
A natural question is, \textit{does this come at the cost of diverse counterfactual examples?}
To answer this question we study the entropy of the distribution that $\tilde{\bm{S}}$ follows by varying $\gamma$ and $n$, and normalize it by $n$ to study the size of typical set of this distribution.
As can be seen in Figure \ref{fig:typical_set}, as long as $\gamma$ is large enough, the average entropy per span $\bar H$ degrades very little from the theoretical maximum, which is $H(p)$, attained when $\gamma=\infty$.
Therefore, to balance between introducing noise in the supporting facts and generating diverse examples, we set $\gamma=1$ in our experiments.

\xhdr{Beta-Bernoulli distribution in dropout} The beta-Bernoulli distribution has been studied in prior work in seeking replacements for the (Bernoulli) dropout mechanism \citep{srivastava2014dropout}. 
\citet{liu2019beta} set $\alpha=\beta$ for the beta distribution in their formulation, which limits the dropout rate to always be 0.5.
\citet{lee2018adaptive} fix $\beta=1$ and vary $\alpha$ to control the sparsity of the result of dropout, which is similar to Beta-\methodname{} when $\gamma=1$.
However, we note that these approaches (as with dropout) are focused more on adding noise to internal representations of neural networks to introduce regularization, while \methodname{} operates directly on the input to ablate different components therein, and thus orthogonal (and potentially complementary) to these approaches.
Further, \methodname{} has the benefit of not having to make any assumption about the model or any change to it during training, which makes it much more widely applicable.

%% file: toytask.tex
In this section, we design a synthetic task of finding a specific subsequence in a character sequence to: a) demonstrate the effectiveness of \methodname{} and Beta-\methodname{} in promoting the performance over a series of problems with different settings, b) analyze the various factors that may affect the efficacy of these approaches, and c) compare it to other counterfactual augmentation techniques like masking on mitigating position bias.

\subsection{Experimental Setup}

\xhdr{\toytask{}} To understand the effectiveness of \methodname{} and Beta-\methodname{} empirically, we designed a synthetic task called \toytask{} where the model is trained to discern that given an animal name $\bm{a}$, \eg, ``cat'', whether a character string contains it as a subsequence (\ie, contains characters in ``cat'' in order, for instance, ``\bfr{a}b\bfg{c}d\bfg{a}fgbijk\bfg{t}m\bfr{a}'') or not (\eg, ``\bfr{a}b\bfr{c}defh\bfr{t}ijk\bfr{a}mn'').
This allows us to easily control the total sequence length $n$, the supporting facts size $m$, as well as easily estimate the supporting fact noise that each \methodname{} variant might introduce.%
\footnote{To generate the synthetic training data of \toytask{}, we first generate sequences consisting of lowercase letters (a to z) that each sequence $\bm{S}$ does not contain the animal name $\bm{a}$; then randomly choose half of these sequences, and replace letters with those in $\bm{a}$ from arbitrary (but not necessarily contiguous) positions in $\bm{S}$ to generate positive examples, and label the rest negative.}

In all of our experiments, we evaluate model performance on a held-out set of 10,000 examples to observe classification error. 
We set sequence length to $n=300$ where each letter is a separate span, and chose positions for the letters in the animal name $\bm{a}$ ($|\bm{a}|=m=3$) uniformly at random in the sequence unless otherwise mentioned.

\xhdr{Model} We employ a three-layer Transformer model \citep{vaswani2017attention} with position embeddings \citep{devlin2019bert} as the sequence encoder implemented with HuggingFace Transformers~\citep{wolf2019huggingface}.  
For each example $(\bm{a}, \bm{S}, y)$, we feed ``[CLS] $\bm{a}$ [SEP] $\bm{S}$ [SEP]'' to the model and then perform binary classifier over the ``[CLS]'' representation to predict $y \in \{0,1\}$, where $y=1$ means $\bm{a}$ appears in $\bm{S}$. 
To investigate the effectiveness of \methodname{}, we apply \methodname{} to $\bm{S}$ first before feeding the resulting sequence into the Transformer classifier.

\subsection{Results and Analysis}

\begin{figure*}[!ht]
    \small
    \centering
    \pgfplotstableread[row sep=\\,col sep=&]{
    	k & base & d1 & do1 & b1 & bo1 & d2 & do2 & b2 & bo2 \\
    	10 & 49.87 & 49.25 & 47.41 & 48.45 & 48.52 & 48.55 & 46.95 & 47.76 & 45.77 \\
20 & 49.71 & 47.56 & 47.60 & 45.31 & 44.43 & 47.39 & 46.81 & 43.52 & 40.33 \\
50 & 47.73 & 44.31 & 43.93 & 40.48 & 35.76 & 41.57 & 39.57 & 40.84 & 30.22 \\
100 & 46.48 & 41.46 & 41.42 & 39.69 & 26.48 & 42.04 & 32.01 & 36.05 & 22.02 \\
200 & 41.89 & 28.42 & 19.35 & 20.34 & 10.71 & 31.88 & 26.24 & 19.25 & 10.43 \\
500 & 38.35 & 20.09 & 16.67 & 11.59 & 6.65 & 29.26 & 22.82 & 12.12 & 5.05 \\
1000 & 30.57 & 18.31 & 15.10 & 6.93 & 2.94 & 27.68 & 25.27 & 7.29 & 2.85 \\
2000 & 19.49 & 16.74 & 14.60 & 3.54 & 1.93 & 26.02 & 24.52 & 2.34 & 1.66 \\
5000 & 9.69 & 14.75 & 13.57 & 1.21 & 0.96 & 23.87 & 22.84 & 1.22 & 0.91 \\
10000 & 5.37 & 13.62 & 11.27 & 0.92 & 0.76 & 23.43 & 21.20 & 0.96 & 0.65 \\
20000 & 2.31 & 12.14 & 13.52 & 0.74 & 0.57 & 19.22 & 20.82 & 0.83 & 0.76 \\
50000 & 1.39 & 11.61 & 9.90 & 0.87 & 0.63 & 25.08 & 25.95 & 0.92 &  \\
    }\dataefficiency
    
    \pgfplotstableread[row sep=\\,col sep=&]{
    	k & base & d1 & do1 & b1 & bo1 & d2 & do2 & b2 & bo2 \\
    	10 & 49.87 & 47.76 & 47.57 & 47.69 & 47.43 & 47.30 & 46.76 & 47.46 &  \\
20 & 47.66 & 46.49 & 46.70 & 46.37 & 46.65 & 46.49 & 46.65 &  &  \\
50 & 49.43 & 46.44 & 46.57 & 46.28 & 46.35 & 46.00 &  &  &  \\
100 & 48.18 & 45.65 & 43.36 & 41.58 & 37.75 & 42.74 &  &  &  \\
200 & 46.53 & 42.49 & 36.90 & 38.98 & 30.42 & 41.63 &  &  &  \\
500 & 46.06 & 26.85 & 21.09 & 21.94 & 11.62 & 34.08 &  &  &  \\
1000 & 43.97 & 23.37 & 19.24 & 11.97 & 7.03 & 30.36 &  &  &  \\
2000 & 38.96 & 18.15 & 17.16 & 6.42 & 2.56 & 30.81 &  &  &  \\
5000 & 15.67 & 15.99 & 14.31 & 1.55 & 1.62 & 27.01 &  &  &  \\
10000 & 9.19 & 14.68 & 14.49 & 1.41 & 1.00 &  &  &  &  \\
20000 & 5.19 & 11.16 & 13.84 & 0.94 & 0.96 &  &  &  &  \\
50000 & 1.92 & 15.73 & 15.76 & 1.00 & 0.99 &  &  &  &  \\
    }\dataefficiencytwo
    
    \pgfplotstableread[row sep=\\,col sep=&]{
    	k & base & d1 & do1 & b1 & bo1 & d2 & do2 & b2 & bo2 \\
    	10 & 49.15 & 49.07 & 49.03 & 49.00 & 49.20 & 49.09 & 49.07 & 49.23 & 49.06 \\
20 & 49.93 & 49.38 & 49.43 & 49.19 & 49.24 & 49.45 & 49.41 & 49.34 & 49.33 \\
50 & 49.36 & 49.00 & 49.01 & 49.00 & 48.90 & 48.94 & 48.92 & 48.90 & 48.49 \\
100 & 49.05 & 49.07 & 49.21 & 48.81 & 48.81 & 49.08 & 49.20 & 48.42 & 47.76 \\
200 & 48.39 & 47.98 & 47.49 & 47.64 & 46.82 & 47.75 & 46.83 & 46.59 & 44.69 \\
500 & 48.37 & 46.28 & 45.56 & 45.22 & 43.69 & 45.61 & 44.83 & 44.48 & 39.46 \\
1000 & 48.18 & 44.78 & 43.92 & 43.58 & 40.97 & 44.72 & 41.37 & 42.17 & 36.89 \\
2000 & 47.49 & 37.98 & 33.75 & 34.56 & 29.29 & 39.05 & 35.12 & 32.02 & 22.18 \\
5000 & 46.25 & 22.70 & 22.46 & 14.55 & 12.83 & 28.59 & 28.29 & 12.49 & 10.23 \\
10000 & 43.48 & 19.86 & 19.68 & 10.58 & 9.26 & 28.46 & 27.15 & 11.77 & 7.34 \\
20000 & 30.45 & 18.08 & 18.24 & 7.68 & 6.34 & 27.03 & 27.13 & 10.53 & 7.83 \\
50000 & 10.97 & 18.67 & 17.70 & 8.88 & 7.59 & 26.98 & 26.82 & 10.41 & 7.44 \\
    }\dataefficiencyten
    
    \pgfplotstableread[row sep=\\,col sep=&]{
    	p & d & do & b & bo \\
    	0 & 31.26 & 31.26 & 31.26 & 31.26 \\
0.05 & 13.68 & 9.79 & 7.90 & 4.98 \\
0.1 & 18.31 & 15.10 & 6.93 & 2.94 \\
0.15 & 20.32 & 18.87 & 7.11 & 2.74 \\
0.2 & 27.68 & 25.27 & 7.29 & 2.85 \\
0.3 & 33.32 & 31.60 & 6.77 & 2.45 \\
0.4 & 35.83 & 37.67 & 6.27 & 2.73 \\
0.5 & 39.32 & 41.96 & 6.54 & 2.86 \\
    }\lblnoise
   
    \pgfplotstableread[row sep=\\,col sep=&]{
    	p & d & b \\
    	0 & 0.00 & 0.00 \\
0.05 & 2.00 & 1.75 \\
0.1 & 3.78 & 4.01 \\
0.15 & 5.75 & 6.42 \\
0.2 & 8.08 & 8.90 \\
0.3 & 12.84 & 13.84 \\
0.4 & 18.33 & 19.83 \\
0.5 & 24.27 & 25.82 \\
    }\lblnoisebar
    
    \pgfplotstableread[row sep=\\,col sep=&]{
    	l & base & d & do & b & bo \\
    	10 & 0.24 & 0.60 & 0.05 & 0.36 & 0.04 \\
20 & 1.95 & 2.82 & 0.22 & 1.97 & 0.14 \\
30 & 4.14 & 5.72 & 1.11 & 4.50 & 0.54 \\
50 & 7.88 & 8.95 & 1.88 & 6.08 & 0.88 \\
100 & 14.56 & 16.55 & 10.97 & 7.19 & 2.41 \\
200 & 23.48 & 16.51 & 12.82 & 9.28 & 3.43 \\
300 & 31.26 & 18.31 & 15.10 & 6.93 & 2.94 \\
500 & 39.54 & 23.80 & 18.52 & 7.84 & 4.53 \\
    }\seqlen

    \pgfplotstableread[row sep=\\,col sep=&]{
    	l & d & b \\
    10 & 11.03 & 10.40 \\
20 & 9.73 & 9.90 \\
30 & 9.25 & 8.89 \\
50 & 7.78 & 7.71 \\
100 & 4.95 & 5.53 \\
200 & 4.41 & 4.60 \\
300 & 3.78 & 4.01 \\
500 & 3.29 & 3.33 \\
    }\seqlenbar
    
    \pgfplotstableread[row sep=\\,col sep=&]{
    	l & base & d & do & b & bo \\
2 & 15.22 & 15.94 & 14.35 & 2.20 & 1.15 \\
3 & 30.49 & 19.80 & 17.48 & 5.35 & 2.51 \\
4 & 36.07 & 22.48 & 19.66 & 12.48 & 8.98 \\
5 & 35.53 & 23.71 & 21.63 & 17.48 & 14.56 \\
6 & 33.15 & 23.60 & 22.75 & 20.92 & 16.53 \\
7 & 35.36 & 26.64 & 25.36 & 21.83 & 19.30 \\
8 & 34.64 & 27.12 & 26.01 & 22.03 & 20.45 \\
9 & 37.68 & 27.27 & 26.96 & 21.75 & 21.14 \\
10 & 32.40 & 27.23 & 26.80 & 22.30 & 22.42 \\
    }\tgtlen

    \pgfplotstableread[row sep=\\,col sep=&]{
    	l & d & b \\
2 & 3.54 & 3.72 \\
3 & 3.64 & 3.75 \\
4 & 4.03 & 4.34 \\
5 & 3.71 & 4.03 \\
6 & 3.70 & 4.02 \\
7 & 4.01 & 4.86 \\
8 & 4.57 & 5.49 \\
9 & 4.91 & 5.68 \\
10 & 5.60 & 6.53 \\
    }\tgtlenbar
    
    \pgfplotstableread{
    l	base	d	b	do	bo
325	34.36	28.69	25.48	28.73	24.39
315	33.79	28.05	22.31	27.36	21.25
310	33.05	25.26	19.13	24.37	17.51
305	33.03	23.30	15.00	21.82	13.58
300	31.77	21.54	11.09	20.00	9.31
295	32.50	18.72	10.85	17.70	9.07
290	32.06	17.14	10.82	15.42	9.03
285	31.94	14.26	10.67	12.82	8.76
280	31.91	12.01	10.18	10.79	8.71
275	32.34	11.68	10.71	11.06	8.98
270	32.99	11.72	11.02	11.01	9.39
265	32.23	11.18	10.26	10.29	9.11
260	32.55	11.76	10.58	11.12	9.45
255	33.18	11.92	10.88	10.91	9.59
250	33.18	12.19	11.17	11.68	10.28
225	34.04	13.92	12.57	13.50	12.48
200	34.05	17.71	16.80	18.09	17.04
    }\testlen
    
    \pgfplotstableread{
    setting	base	drop	mask
    1    31.77	21.54	27.12
    2    31.77	11.09	26.45
    }\dropvsmask
    
    \pgfplotstableread{
    setting base    drop    bdrop   mask
    1	49.84	35.53	31.23   46.45
2	32.61	32.06	25.31   30.37
    }\positionzeroshot
    
    \pgfplotstableread[row sep=\\,col sep=&]{
    l & base & d & do & b & bo \\
100 & 49.87 & 45.52 & 43.46 & 49.86 & 49.87  \\
1000 & 40.99 & 40.2 & 40.59 & 40.09 & 43.46  \\
2000 & 35.67 & 33.2 & 31.19 & 28.8 & 22.93  \\
3000 & 33.6 & 26.5 & 27.06 & 20.37 & 16.31  \\
4000 & 35.97 & 31.53 & 33.62 & 22.97 & 16.06  \\
5000 & 34.59 & 28.97 & 25.94 & 16.82 & 17.37  \\
6000 & 33.65 & 24.83 & 22.29 & 10.13 & 11.28  \\
7000 & 34.46 & 27.68 & 25.9 & 10.21 & 13.04  \\
8000 & 33.99 & 22.67 & 22.92 & 7.91 & 9.13  \\
9000 & 35.34 & 24.07 & 20.64 & 10.7 & 10.56  \\
10000 & 32.55 & 24.38 & 26.13 & 12.19 & 11.97  \\
11000 & 31.89 & 32.38 & 23.18 & 10.03 & 6.25  \\
12000 & 32.11 & 23.03 & 21.31 & 8.38 & 5.59  \\
13000 & 32.69 & 25.56 & 21.43 & 8.03 & 6.38  \\
14000 & 32.24 & 22.13 & 20.13 & 6.1 & 5.83  \\
15000 & 31.95 & 23.14 & 21.11 & 7.6 & 7.54  \\
16000 & 34.2 & 26.06 & 24.34 & 8.59 & 5.18  \\
17000 & 31.85 & 25.06 & 19.83 & 7.13 & 5.18  \\
18000 & 33.15 & 24.99 & 22.06 & 10.82 & 6.57  \\
19000 & 31.65 & 23.63 & 20.7 & 6.53 & 5.58  \\
20000 & 32.26 & 27.21 & 21.53 & 7.53 & 5.11  \\
21000 & 33.58 & 24.95 & 19.59 & 11.83 & 5.24  \\
22000 & 32.91 & 23.15 & 23.03 & 6.39 & 6.62  \\
23000 & 31.91 & 26.58 & 20.48 & 6.14 & 5.9  \\
24000 & 33.41 & 21.78 & 20.76 & 7.93 & 4.09  \\
25000 & 32.34 & 24.79 & 21.19 & 6.53 & 4.33  \\
26000 & 32.75 & 28.18 & 24.09 & 14.87 & 4.22  \\
27000 & 32.16 & 26.25 & 20.46 & 6.19 & 4.33  \\
28000 & 31.68 & 23.7 & 20.31 & 9.54 & 4.32  \\
29000 & 31.56 & 24.91 & 21.08 & 9.7 & 6.01  \\
30000 & 32.03 & 26.58 & 21.96 & 7.69 & 3.87  \\
31000 & 32.14 & 24.64 & 19.81 & 8.25 & 5.36  \\
32000 & 32.19 & 25.91 & 20.5 & 8.91 & 4.46  \\
33000 & 32.58 & 29.35 & 21.71 & 6.14 & 3.5  \\
34000 & 32.16 & 24.31 & 19.58 & 7.01 & 3.75  \\
35000 & 31.93 & 29.79 & 19.94 & 11.15 & 3.3  \\
36000 & 32.23 & 25.58 & 20.4 & 13.14 & 3.45  \\
37000 & 31.85 & 22.71 & 18.76 & 6.99 & 3.18  \\
38000 & 32.7 & 23.3 & 19.21 & 9.47 & 3.72  \\
39000 & 31.9 & 24.36 & 22.79 & 8.86 & 4.02  \\
40000 & 30.99 & 26.51 & 19.19 & 9.34 & 2.92  \\
41000 & 31.47 & 28.17 & 20.77 & 12.97 & 6.46  \\
42000 & 31.54 & 26.7 & 21.16 & 8.24 & 3.45  \\
43000 & 33.56 & 26.22 & 19.9 & 8.12 & 4.51  \\
44000 & 31.16 & 31.17 & 20.56 & 13.43 & 3.3  \\
45000 & 31.03 & 33.43 & 19.91 & 10.36 & 3.81  \\
46000 & 31.14 & 25.8 & 19.13 & 10.18 & 2.9  \\
47000 & 31.93 & 25.99 & 21.23 & 10.73 & 3.3  \\
48000 & 31.78 & 29.26 & 18.64 & 11.94 & 2.62  \\
49000 & 31.94 & 28.69 & 21.95 & 13.55 & 2.75  \\
50000 & 31.89 & 26.26 & 19.55 & 10.27 & 2.78  \\
    }\training
    
    \pgfplotstableread[row sep=\\,col sep=&]{
    l & base & d & do & b & bo \\
100 & 49.87 & 45.52 & 43.46 & 49.86 & 49.87  \\
5000 & 34.59 & 28.97 & 25.94 & 16.82 & 17.37  \\
10000 & 32.55 & 24.38 & 26.13 & 12.19 & 11.97  \\
15000 & 31.95 & 23.14 & 21.11 & 7.6 & 7.54  \\
20000 & 32.26 & 27.21 & 21.53 & 7.53 & 5.11  \\
25000 & 32.34 & 24.79 & 21.19 & 6.53 & 4.33  \\
30000 & 32.03 & 26.58 & 21.96 & 7.69 & 3.87  \\
35000 & 31.93 & 29.79 & 19.94 & 11.15 & 3.3  \\
40000 & 30.99 & 26.51 & 19.19 & 9.34 & 2.92  \\
45000 & 31.03 & 33.43 & 19.91 & 10.36 & 3.81  \\
50000 & 31.89 & 26.26 & 19.55 & 10.27 & 2.78  \\
    }\trainingmarkers
    
    \begin{tikzpicture}[every plot/.append style={thick},font=\footnotesize,every axis/.append style={scale only axis, axis on top,xmode=log,ymode=log,ytick={0.01, 0.03, 0.1, 0.3, 1, 3, 10, 30},yticklabels={0.01, 0.03, 0.1, 0.3, 1, 3, 10, 30},}]
    \begin{groupplot}[group style={group size= 4 by 1},height=2cm,width=3.25cm]
    \nextgroupplot[title={(a) Data efficiency},xlabel={\# Training examples},ylabel={Error/Noise (\%)},legend to name=lgd,legend style={legend columns=5,nodes={scale=0.7, transform shape}}]
        \addplot[color=gray,mark=x] plot table[x=k, y=base]{\dataefficiency}; \addlegendentry{Baseline}
        \addplot[color=red,mark=*] plot table[x=k, y=d1]{\dataefficiency}; \addlegendentry{\methodname{}}
        \addplot[dashed,color=red,mark=o,mark options=solid] plot table[x=k, y=do1]{\dataefficiency}; \addlegendentry{\methodname{} (noise-free)}
        \addplot[color=blue,mark=square*] plot table[x=k, y=b1]{\dataefficiency}; \addlegendentry{Beta-\methodname{} }
        \addplot[dashed,color=blue,mark=square,mark options=solid] plot table[x=k, y=bo1]{\dataefficiency};\addlegendentry{Beta-\methodname{} (noise-free)}
        
        \coordinate (top) at (rel axis cs:0,1);%
    \nextgroupplot[title={(b) Noise in supporting facts},xmode=linear,legend style={legend columns=1,nodes={scale=0.5, transform shape}},ymax=1000,xlabel={Drop rate $p$},xtick={0, 0.1, 0.2, 0.3, 0.4, 0.5}]
        \addplot[ybar,ybar legend,bar width=0.1cm,bar shift=-0.07cm,color=orange]  table[x=p, y=d]{\lblnoisebar};\addlegendentry{Noise, \methodname{}}
        \addplot[ybar,ybar legend,bar width=0.1cm,bar shift=0.07cm,color=green] plot table[x=p, y=b]{\lblnoisebar};\addlegendentry{Noise, Beta-\methodname{}}
        \addplot[color=red,mark=*] plot table[x=p, y=d]{\lblnoise}; 
        \addplot[dashed,color=red,mark=o,mark options=solid] plot table[x=p, y=do]{\lblnoise}; 
        \addplot[color=blue,mark=square*] plot table[x=p, y=b]{\lblnoise};
        \addplot[dashed,color=blue,mark=square,mark options=solid] plot table[x=p, y=bo]{\lblnoise};
    \nextgroupplot[title={(c) Varying sequence length},log origin=infty,legend style={legend columns=1,nodes={scale=0.5, transform shape}},ymax=20000,xtick={10,30,100,300},xticklabels={10,30,100,300},xlabel={Train/test sequence length},ytick={0.1,1,10},yticklabels={0.1,1,10}]
        \addplot[ybar,ybar legend,bar width=0.1cm,bar shift=-0.07cm,color=orange] plot table[x=l, y=d]{\seqlenbar};\addlegendentry{Noise, \methodname{}}
        \addplot[ybar,ybar legend,bar width=0.1cm,bar shift=0.07cm,color=green] plot table[x=l, y=b]{\seqlenbar};\addlegendentry{Noise, Beta-\methodname{}}
        \addplot[color=gray,mark=x] plot table[x=l, y=base]{\seqlen}; 
        \addplot[color=red,mark=*] plot table[x=l, y=d]{\seqlen}; 
        \addplot[dashed,color=red,mark=o,mark options=solid] plot table[x=l, y=do]{\seqlen}; 
        \addplot[color=blue,mark=square*] plot table[x=l, y=b]{\seqlen};
        \addplot[dashed,color=blue,mark=square,mark options=solid] plot table[x=l, y=bo]{\seqlen}; 
    \nextgroupplot[title={(d) Varying supporting fact size},log origin=infty,legend style={legend columns=1,nodes={scale=0.5, transform shape}},ymax=3000,xmode=linear,xtick={2,4,6,8,10},xticklabels={2,4,6,8,10},xlabel={Supporting fact size},ytick={0.1,0.3,1,3,10,30},yticklabels={0.1,0.3,1,3,10,30}]
        \addplot[ybar,ybar legend,bar width=0.1cm,bar shift=-0.07cm,color=orange] plot table[x=l, y=d]{\tgtlenbar};\addlegendentry{Noise, \methodname{}}
        \addplot[ybar,ybar legend,bar width=0.1cm,bar shift=0.07cm,color=green] plot table[x=l, y=b]{\tgtlenbar};\addlegendentry{Noise, Beta-\methodname{}}
        \addplot[color=gray,mark=x] plot table[x=l, y=base]{\tgtlen}; 
        \addplot[color=red,mark=*] plot table[x=l, y=d]{\tgtlen}; 
        \addplot[dashed,color=red,mark=o,mark options=solid] plot table[x=l, y=do]{\tgtlen}; 
        \addplot[color=blue,mark=square*] plot table[x=l, y=b]{\tgtlen};
        \addplot[dashed,color=blue,mark=square,mark options=solid] plot table[x=l, y=bo]{\tgtlen}; 
        \coordinate (bot) at (rel axis cs:1,0);
    \end{groupplot}
    
    \path (top|-current bounding box.north)--
      coordinate(legendpos)
      (bot|-current bounding box.north);
    \node[right=1em,inner sep=1pt, anchor=south] at(legendpos) {\pgfplotslegendfromname{lgd}};
  
    \end{tikzpicture}
    
    \begin{tikzpicture}[every plot/.append style={thick},font=\footnotesize,every axis/.append style={scale only axis, axis on top,ymin=0}]
    \begin{groupplot}[group style={group size= 4 by 1},height=2cm,width=3.25cm]
    \nextgroupplot[title={(e) OOD length generalization},log origin=infty,legend style={legend columns=1,nodes={scale=0.7, transform shape}},ylabel={Error(\%)},xmode=linear,xtick={200,225,250,275,300,325},xlabel={Test sequence length}]
        \addplot[color=gray,mark=x] plot table[x=l, y=base]{\testlen}; 
        \addplot[color=red,mark=*] plot table[x=l, y=d]{\testlen}; 
        \addplot[dashed,color=red,mark=o,mark options=solid] plot table[x=l, y=do]{\testlen}; 
        \addplot[color=blue,mark=square*] plot table[x=l, y=b]{\testlen};
        \addplot[dashed,color=blue,mark=square,mark options=solid] plot table[x=l, y=bo]{\testlen}; 
        
    \nextgroupplot[title={(f) \methodname{} vs \spanmask{}},log origin=infty,legend style={legend columns=1,nodes={scale=0.7, transform shape}},xmode=linear,xlabel={Drop/mask distribution},xtick={1, 2},xticklabels={Bernoulli, Beta-Bernoulli},xmin=0.5,xmax=2.5,ymax=45]
        \addplot[ybar,color=orange,mark=diamond*,bar width=0] plot table[x=setting, y=mask]{\dropvsmask}; \addlegendentry{\spanmask{}}
        \addplot[ybar,color=gray,mark=x,bar width=0,xshift=-0.3cm] plot table[x=setting, y=base]{\dropvsmask}; 
        \addplot[ybar,color=red,mark=*,bar width=0,xshift=0.3cm] plot table[x=setting, y=drop]{\dropvsmask}; 
    \nextgroupplot[title={(g) Position\&0-shot generalization},log origin=infty,legend style={legend columns=1,nodes={scale=0.5, transform shape}},xmode=linear,xlabel={Setting},xtick={1, 2},xticklabels={Fixed, First 100},xmin=0.5,xmax=2.5, ymax=62]
        \addplot[ybar,color=orange,mark=diamond*,bar width=0,xshift=-0.1cm] plot table[x=setting, y=mask]{\positionzeroshot}; \addlegendentry{\spanmask{}}
        \addplot[ybar,color=gray,mark=x,bar width=0,xshift=-0.3cm] plot table[x=setting, y=base]{\positionzeroshot}; 
        \addplot[ybar,color=red,mark=*,bar width=0,xshift=0.1cm] plot table[x=setting, y=drop]{\positionzeroshot}; 
        \addplot[ybar,color=blue,mark=square*,bar width=0,xshift=0.3cm] plot table[x=setting, y=bdrop]{\positionzeroshot}; 
    \nextgroupplot[title={(h) Dev error during training},log origin=infty,legend style={legend columns=1,nodes={scale=0.5, transform shape}},xmode=linear,ymode=log,xlabel={Training steps},ytick={3,10,30},yticklabels={3,10,30},xtick={0, 25000, 50000},xticklabels={0,2.5,\qquad 5\ $\cdot$10\textsuperscript{4}},scaled x ticks=false]
        \addplot[color=gray] plot table[x=l, y=base]{\training}; 
        \addplot[only marks,color=gray,mark=x] plot table[x=l, y=base]{\trainingmarkers};
        \addplot[color=red] plot table[x=l, y=d]{\training}; 
        \addplot[only marks,color=red,mark=*] plot table[x=l, y=d]{\trainingmarkers};
        \addplot[dashed,color=red] plot table[x=l, y=do]{\training}; 
        \addplot[only marks,color=red,mark=o,mark options=solid] plot table[x=l, y=do]{\trainingmarkers};
        \addplot[color=blue] plot table[x=l, y=b]{\training};
        \addplot[only marks,color=blue,mark=square*] plot table[x=l, y=b]{\trainingmarkers};
        \addplot[dashed,color=blue] plot table[x=l, y=bo]{\training}; 
        \addplot[only marks,color=blue,mark=square] plot table[x=l, y=bo]{\trainingmarkers};
    \end{groupplot}
    \end{tikzpicture}
    \vspace{-1em}
    \caption{Experimental results of \methodname{} variants and \spanmask{} on the \toytask{} synthetic tasks.}
    \label{fig:toy_task}
\end{figure*}
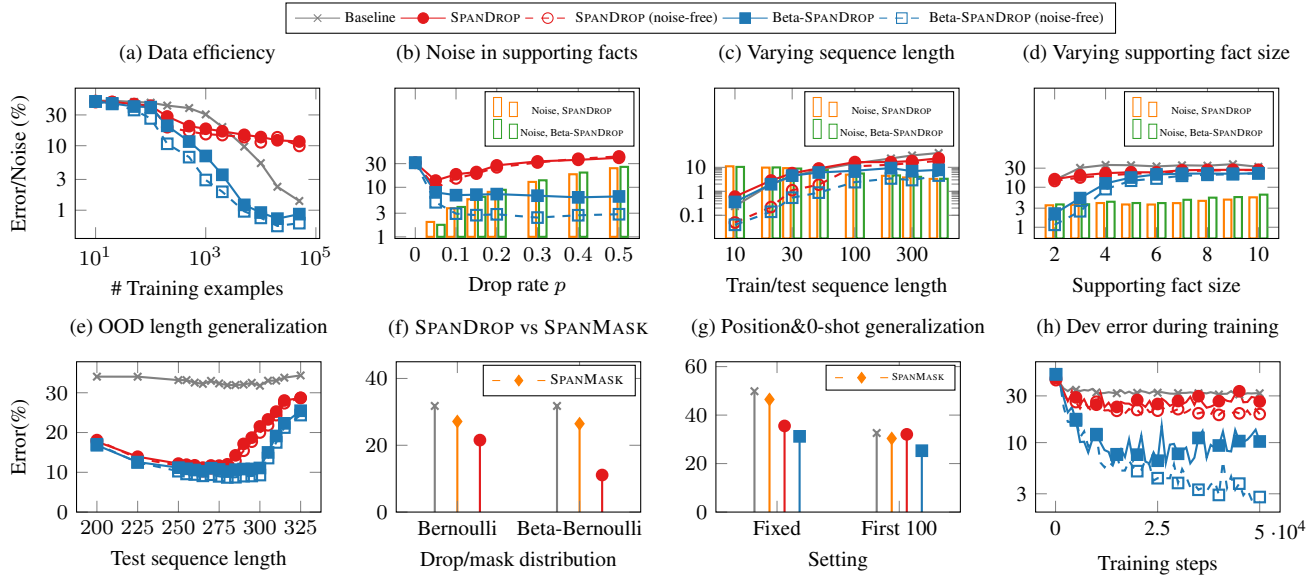

In each experiment, we compare \methodname{} and Beta-\methodname{} at the same drop ratio $p$.
And we further use rejection sampling to remove examples that do not preserve the desired supporting facts to understand the effect of supporting fact noise.

\xhdr{Data efficiency} We begin by analyzing the contribution of \methodname{} and Beta-\methodname{} to improving the sample efficiency of the baseline model.
To achieve this goal, we vary the training set size from 10 to 50,000 and observe the prediction error on the held-out set.
We observe from the results in Figure \ref{fig:toy_task}(a) that: 1) Both \methodname{} and Beta-\methodname{} significantly improve data efficiency in low-data settings.
For instance, when trained on only 200 training examples, \methodname{} variants can achieve the generalization performance of the baseline model trained on 5x to even 20x data.
2) Removing supporting fact noise typically improves data efficiency further by about 2x.
This indicates it is helpful not to drop spans in $S_\mathrm{sup}$ during training when possible, so that the model is always trained with true counterfactual examples rather than sometimes noisy ones.
3) Beta-\methodname{} consistently improves upon the baseline model even when data is abundant.
This is likely due to the difficulty of the task when $n=300$ and $m=3$. 
Similar to many real-world tasks, the task remains underspecified even when the generalization error is already very low, thanks to the large amount of training data available.
4) \methodname{} introduces inconsistent training objective with the original training set, which leads to performance deterioration when there is sufficient training data, which is consistent with our theoretical observation.

\xhdr{Effect of supporting fact noise} Since \methodname{} introduces noise in the supporting facts (albeit with a low probability), it is natural to ask if such noise is negatively correlated with model performance.
We study this by varying the drop ratio $p$ from $\{0.05, 0.1, 0.15, 0.2, 0.3, 0.4, 0.5\}$ on fixed training sets of size 1,000, and observe the resulting model performance and supporting fact error.
As can be seen in Figure \ref{fig:toy_task}(b), supporting fact noise increases rapidly as $p$ grows.\footnote{Note that the noise in our experiments are lower than what would be predicted by theory, because in practice the initial sequence $\bm{S}$ might already contain parts of $\bm{a}$ before it is inserted. This creates redundant sets of supporting facts for this task and reduces supporting fact noise especially when $n$ is large.}
However, we note that although the performance of \methodname{} deteriorates as $p$ increases, that of Beta-\methodname{} stays relatively stable.
Inspecting these results more closely, we find that even the performance of the noise-free variants follow a similar trend, which should not be affected by supporting fact noise.

\xhdr{Effect of sequence and supporting fact lengths}
In our theoretical analysis, sequence length $n$ determines regularization strength and the drift in sequence length distribution, and supporting fact size $m$ determines the probability that the counterfactual example retains the correct supervision signal.
To study their effects empirically, we conduct three separate sets of experiments: 1) training and testing the model on varying sequence lengths \{10, 20, 30, 50, 100, 200, 300, 500\}; 2) varying supporting fact size between 2 and 10; and 3) testing the model trained on $n=300$ on test sets of different lengths.

As can be seen from Figures \ref{fig:toy_task}(c) and \ref{fig:toy_task}(e), our experimental results seem to well supporting this hypothesis that the gap between the two \methodname{} variants has a lot to do with the discrepancy in length distributions.
Specifically, in Figure \ref{fig:toy_task}(c), while the performance of both \methodname{} variants deteriorates as $n$ grows and the task becomes more challenging and underspecified, \methodname{} deteriorates at a faster speed even when we remove the effect of supporting fact noise.
On the other hand, we see in Figure \ref{fig:toy_task}(e) that \methodname{} performance peak around sequences of length 270 ($=n(1-p)$=300$\times$(1-0.1)) before rapidly deteriorating, while Beta-\methodname{} is unaffected until test sequence length exceeds that of all examples seen during training.

In Figure \ref{fig:toy_task}(d), we can also see that as predicted by Remarks 1 and 2, Beta-\methodname{} is better at preserving all supporting facts than \methodname{} agnostic of their positions, which translates to superior performance.

\xhdr{Mitigating position bias} Besides \methodname{}, replacement-based techniques like masking can also be applied to construct counterfactual examples, where elements in the sequence are replaced by a special symbol that is not used at test time.
We implement \spanmask{} in the same way as \methodname{} except spans are replaced rather than removed when the sampled ``drop mask'' $\delta_i$ is 0.
We first inspect whether \spanmask{} benefits from the same beta-Bernoulli distribution we use in \methodname{}.
As can be seen in Figure \ref{fig:toy_task}(f), the gain from switching to a beta-Bernoulli distribution provides negligible benefit to \spanmask{}, which does not alter the sequence length of the input to begin with.
We also see that \spanmask{} results in significantly higher error than both \methodname{} and Beta-\methodname{} in this setting.

We further experiment with introducing position bias into the training data (but not the test data) to test whether these method help the model generalize to an unseen setting.
Specifically, instead of selecting the position for the characters in $\bm{a}$ uniformly at random, we train the model with a ``fixed position'' dataset where they always occur at indices (10, 110, 210), and a ``first 100'' dataset where they are uniformly distributed among the first 100 letters.
As can be seen in Figure \ref{fig:toy_task}(g), both the baseline and \spanmask{} models overfit to the position bias in the ``fixed'' setting, while \methodname{} techniques significantly reduce zero-shot generalization error.
In the ``first 100'' setting, Beta-\methodname{} consistently outperforms its Bernoulli counterpart and \spanmask{} at improving the performance of the baseline model as well, indicating that \methodname{} variants are effective at reducing the position bias of the model.

\xhdr{Impact on convergence speed} Regularization is commonly shown to slow down model convergence, resulting in the need for much longer training time for marginal improvements in performance.
We show in Figure \ref{fig:toy_task}(h), however, that \methodname{} approaches do not seem to suffer from this issue on the synthetic task.
In the contrary, both approaches help the model generalize significantly better within the same amount of training time.

%% file: experiments.tex
To examine the efficacy of the proposed \methodname{} techniques on realistic data, we conduct experiments on four NLP datasets that represent a variety of tasks.
We focus on showing the effect of \methodname{} instead of pursuing the state-of-the-art in these experiments.

\subsection{Setup and Main Results}
 \begin{table*}[]
\small
    \centering
    \subfigure[HotpotQA dev]{
    \begin{tabular}{lccc}
        \toprule
        Model & Ans \fone{} & Sup \fone{} & Joint \fone{} \\
        \midrule
        RoBERTa-base & 73.5 & 83.4  & 63.5\\
        Longformer-base & 74.3 & 84.4  & 64.4\\
        SAE BERT-base & 73.6 & 84.6 & 65.0 \\
        \midrule
        \textit{Our implementation} \\
        ELECTRA-base & 74.2 & 86.3 & 66.2\\
        \quad + \methodname{} &74.7 & 86.7 & 66.8 \\
        \quad + Beta-\methodname{} &74.7 & 86.9 & 67.1 \\
        \bottomrule
    \end{tabular}
    }
    \subfigure[MultiRC dev/test]{
    \begin{tabular}{lcc}
        \toprule
        Model & EM & \fone{} \\
        \midrule
        BERT-base & 26.6/24.1 & 71.8/70.1 \\
        RoBERTa-base & 38.7/\phantom{0}---\phantom{0} & 77.1/\phantom{0}---\phantom{0}\\
        REPT RoBERTa-base & 40.4/\phantom{0}---\phantom{0} & 80.0/\phantom{0}---\phantom{0} \\
        \midrule
        \textit{Our implementation} \\
        ELECTRA-base & 40.1/39.1 & 80.4/78.2  \\
        \quad + \methodname{} & 42.3/39.9 & 81.7/78.5 \\
        \quad + Beta-\methodname{} & 44.8/41.1 & 81.6/79.8 \\
        \bottomrule
    \end{tabular}
    }
    \subfigure[DocRED dev]{
    \begin{tabular}{lccc}
        \toprule
        Model & Ign \fone{} & RE \fone{} & Evi \fone{} \\
        \midrule
        E2GRE BERT-base & 55.2 & 58.7 & 47.1 \\
        ATLOP BERT-base & 59.2 & 61.1 & ---\\
        SSAN BERT-base & 57.0 & 59.2 & --- \\
        \midrule
        \textit{Our implementation} \\
        ELECTRA-base & 59.6 & 61.6 & 50.8 \\
        \quad + \methodname{} & 59.9 & 61.9 & 51.2 \\
        \quad + Beta-\methodname{} & 60.1 & 62.1 & 51.2 \\
        \bottomrule
    \end{tabular}
    }
    \subfigure[SQuAD dev]{
    \begin{tabular}{lcc}
        \toprule
        Model & EM & \fone{} \\
        \midrule
        RoBERTa-base & --- & 90.6 \\
        ELECTRA-base & 84.5 & 90.8 \\
        XLNet-large & 89.7 & 95.1 \\
        \midrule
        \textit{Our implementation} \\
        ELECTRA-base & 86.6 & 92.4 \\
        \quad + \methodname{} \emph{w/ adaptive spans} & 87.4 & 92.9 \\
        \quad + Beta-\methodname{} \emph{w/ adaptive spans} & 87.3 & 92.8 \\
        \bottomrule
    \end{tabular}
    }
    \vspace{-1em}
    \caption{Main results on four natural language processing datasets.}
    \label{tab:main-results-dev}
\end{table*}

\xhdr{Datasets} 
We use four natural language processing datasets: \textbf{SQuAD} 1.1 \citep{rajpurkar2016squad}, where models answer questions on a paragraph of text from Wikipedia;
 \textbf{MultiRC} \citep{MultiRC2018}, which is a multi-choice reading comprehension task in which questions can only be answered by taking into account information from \textit{multiple sentences};
 \textbf{HotpotQA} \citep{yang2018hotpotqa}, which requires models to perform multi-hop reasoning over multiple Wikipedia pages to answer questions; and \textbf{DocRED}~\citep{yao2019docred}, which is a \textit{document-level} data set for relation extraction.

For the SQuAD dataset, we define spans as collections of one or more consecutive tokens to show that \methodname{} can be applied to different granularities.
Guided by our theoretical analysis, we also experiment with an \emph{adaptive span segmentation} approach to reduce the number of supporting facts ($m$) by combining all $N$-gram overlaps between the question and context into a single span ($N\ge 2$).
For the rest three datasets, we define spans to be sentences since  supporting facts are provided at sentence level.
For all of these tasks, we report standard exact match (EM) and \fone{} metrics where applicable, for which higher scores are better.
We refer the reader to the appendix for details about the statistics and metrics of these datasets.

\xhdr{Model} We build our models based on ELECTRA~\citep{clark2019electra}, since it is shown to perform well across a range of NLP tasks recently.
We introduce randomly initialized task-specific parameters designed for each task following prior work on each dataset, and finetune these models on each dataset to report results.
We refer the reader to the appendix for training details and hyperparameter settings.

\xhdr{Main Results} %
We first present the performance of our implemented models and their combination with \methodname{} variants on the four natural language processing tasks.
We also include results from representative prior work on each dataset for reference (detailed in the appendix), and summarize the results in Table \ref{tab:main-results-dev}.
We observe that: 
1) our implemented models achieve competitive and sometimes significantly better performance (in the cases of HotpotQA, SQuAD, and DocRED) compared to published results, especially considering that we do not tailor our models to each task too much;
2) \methodname{} improves the performance over these models even when the training set is large and that the model is already performing well;
3) Models trained with Beta-\methodname{} consistently perform better or equally well with their \methodname{} counterparts across all datasets, demonstrating that our observations on the synthetic datasets generalize well to real-world ones. 
We note that the performance gains on real-world data is less significant, which likely results from the fact that non-supporting spans in the synthetic task are independent from each other, which is not the case in natural language data.

\subsection{Analysis}
To understand whether the properties of \methodname{} and Beta-\methodname{} we observe on \toytask{} generalize to real-world data, we further perform a set of analysis experiments on SQuAD.
Specifically, we are interested in studying the effect of the amount of training data, the span drop ratio $p$, and the choice of span size on performance.

\begin{figure}
    \centering
    \pgfplotstableread[row sep=crcr]{
    pct base drop beta \\
    0.1 39.41 47.48 46.62\\
    0.2 54.97 59.29 60.87 \\
    0.5 67.04 68.78 70.29 \\
    1 74.17 75.12 75.22 \\
    2 78.48 79.12 79.36 \\
    5 83.44 83.73 83.48\\
    10 86.16 86.91 86.60\\
    20 88.48 89.09 89.01 \\
    50 90.92 90.52 91.20\\
    100 92.44 92.61 92.62 \\
    }\squaddataefficiency
    \pgfplotstableread[row sep=crcr]{
    pct drop beta \\
    0 92.49 92.49\\
    0.03 92.55 92.46\\
    0.05 92.59 92.51\\
    0.1 92.55 92.66 \\
    0.15 92.61 92.57\\
    0.2 92.53 92.59 \\
    0.3 92.09 92.50 \\
    0.4 91.42 92.42\\
    }\squaddropratio
    \pgfplotstableread[row sep=crcr, col sep=&]{
    pct & drop & beta & addrop & adbeta \\
    .25 & & & 92.9 & 92.8 \\
    1 & 92.55 & 92.66 & & \\
    2 & 92.58 & 92.56 & & \\
    4 & 92.60 & 92.56 & & \\
    8 & 92.57 & 92.55 & & \\
    16 & 92.50 & 92.40 & & \\
    32 & 92.39 & 92.44 & & \\
    64 & 92.45 & 92.42 & & \\
    }\squadspansize
    \begin{tikzpicture}[font=\footnotesize,every axis/.append style={scale only axis, axis on top}]
        \begin{groupplot}[group style={group size= 3 by 1},height=3cm,width=2cm]
        \nextgroupplot[xlabel={\% of training set used},ylabel={Dev \fone{} (\%)}, width=1.6cm, height=1.6cm, bar width=0.005cm,legend cell align=center,legend style={anchor=south,at={(2,1.08)},legend columns=3,nodes={scale=0.67, transform shape}},xmode=log,xtick={0.1,1,10,100},xticklabels={0.1,1,10,100}]
            \addplot[color=gray, mark=x] plot table[x=pct, y=base]{\squaddataefficiency}; \addlegendentry{Baseline}
            \addplot[color=red, mark=*] plot table[x=pct, y=drop]{\squaddataefficiency}; \addlegendentry{\methodname{}}
            \addplot[color=blue, mark=square*] plot table[x=pct, y=beta]{\squaddataefficiency};\addlegendentry{Beta-\methodname{}} 
        \nextgroupplot[xlabel={Drop ratio $p$}, width=1.6cm, height=1.6cm, bar width=0.005cm,legend cell align=center,legend style={anchor=south,at={(0.45,1.07)},legend columns=3,nodes={scale=0.67, transform shape}},ymin=91,ymax=93,ytick={91,92,93}]
            \addplot[color=red, mark=*] plot table[x=pct, y=drop]{\squaddropratio}; 
            \addplot[color=blue, mark=square*] plot table[x=pct, y=beta]{\squaddropratio}; 
        \nextgroupplot[xlabel={Span size}, width=1.6cm, height=1.6cm, bar width=0.005cm,legend cell align=center,legend style={anchor=south,at={(0.45,1.07)},legend columns=3,nodes={scale=0.67, transform shape}},xmode=log,xtick={.25,1,4,16,64},xticklabels={ad,1,4,16,64},ymin=91,ymax=93,ytick={91,92,93}]
            \addplot[color=red, mark=*] plot table[x=pct, y=drop]{\squadspansize}; 
            \addplot[color=blue, mark=square*] plot table[x=pct, y=beta]{\squadspansize}; 
            \addplot[color=red, mark=*] plot table[x=pct, y=addrop]{\squadspansize}; 
            \addplot[color=blue, mark=square*] plot table[x=pct, y=adbeta]{\squadspansize}; 
        \end{groupplot}
    \end{tikzpicture}
    \vspace{-.3em}
    \caption{Effect analysis of training data size, drop ratio and span size on performance of models trained with \methodname{} and Beta-\methodname{} over SQuAD}
    \vspace{-.7em}
    \label{fig:squad_analysis}
\end{figure}
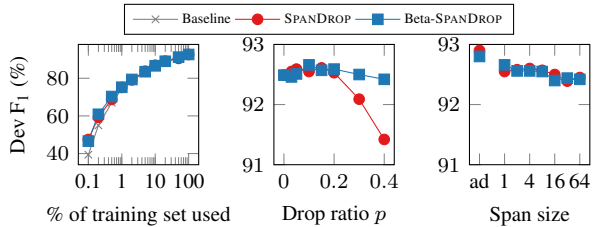

\xhdr{Effect of low data} 
To understand \methodname{}'s regularizing effect when training data is scarce, we study the model's generalization performance when training on only 0.1\% of the training data (around 100 examples) to using the entire training set (around 88k examples).
As shown in Figure \ref{fig:squad_analysis} (left), both \methodname{} and Beta-\methodname{} significantly improve model performance when the amount of training data is extremely low.
As the amount of training data increases, this gap slowly closes but remains consistently positive.
When all training data is used, the gap is still sufficient to separate top-2 performing systems on this dataset.

\xhdr{Impact of drop ratio} 
We compare \methodname{} and Beta-\methodname{} by controlling how likely each span is dropped on average (drop ratio $p$).
Recall from our experiments on \toytasksmall{} that larger $p$ will result in distribution drift from the original training set for \methodname{} but not Beta-\methodname{}, thus the performance of the former deteriorates as $p$ increases while the latter is virtually not affected.
As can be seen in Figure \ref{fig:squad_analysis} (middle), our observation on real-world data is consistent with this theoretical prediction, and indicate that Beta-\methodname{} is a better technique for data augmentation should one want to increase sequence diversity by setting $p$ to a larger value.

\xhdr{Impact of span selection} 
We train the model with \methodname{} on SQuAD with varying fixed span sizes of \{1, 2, 4, 8, 16, 32, 64\} tokens per span, as well as task-motivated \emph{adaptive} span sizes, to understand the effect of this hyperparameter.
We observe in Figure \ref{fig:squad_analysis} (right) that as span size grows, the generalization performance of the model first holds roughly constant, then slowly deteriorates as span size grows too large.
Adaptive spans (``ad'') notably outperforms all fixed span sizes, suggesting that while larger fixed span sizes reduce $m$ and thus noise in supporting facts, the resulting reduction in $n$ hinders regularization strength and thus hurting generalization.
This observation is consistent with that on our synthetic data as well as what would be predicted by our theoretical analysis.
This also suggests that while \methodname{} works with arbitrary span sizes, the optimal choice of spans for different tasks warrants further investigation, which we leave to future work.

%% file: relatedwork.tex
\xhdr{Long Sequence Inference} Many applications require the prediction/inference over long sequences, such as multi-hop reading comprehension~\citep{yang2018hotpotqa,welbl2018constructing}, long document summarization~\citep{huang2021efficient}, document-level information extraction~\citep{yao2019docred} in natural language processing, long sequence time-series prediction ~\citep{zhou2021informer}, promoter region and chromatin-profile prediction in DNA sequence~\citep{oubounyt2019deepromoter,zhou2015predicting} in Genomics etc, where not all elements in the long sequence contribute equally to the desired output. 
Aside from approaches that attempt to approximate all pair-wise interactions between elements in a sequence, more recent work has also investigated compressing long sequences into shorter ones to distill the information therein for prediction or representation learning \citep{Rae2020Compressive,goyal2020power,kim-cho-2021-length}.

\xhdr{Sequence Data Augmentation}
Data augmentation is an effective common technique for underspecified tasks like long sequence inference.
\citet{feng2021survey} propose to group common data augmentation techniques in natural language processing into three categories:
1) rule-based methods~\citep{zhang2015character,wei2019eda,csahin2018data}, which apply a set of predefined operations over the raw input, such as removing, adding, shuffling and replacement; 
2) example mixup-based methods~\citep{guo2019augmenting, guo2020nonlinear,chen2020mixtext, jindal2020augmenting}, which,  inspired from Mixup in computer vision~\citep{zhang2018mixup},  perform interpolation between continuous features like word embeddings and sentence embeddings; 
3) model-based methods~\citep{xie2020unsupervised,sennrich2016improving}, which use trained models to generate new examples \citep[\eg, back translation ][]{xie2020unsupervised}. %

Most existing rule-based data augmentation methods operate at the token/word level~\citep{feng2021survey}, such as word shuffle/replacement/addition~\citep{wei2019eda}. 
Shuffle-based techniques are less applicable when order information is crucial in the raw data \citep[\eg, in natural language]{lan2019albert}. 
\citet{clark2019electra} and \citet{lewis-etal-2020-bart} have recently shown that well designed word-level augmentation can also lead to improved pretrained models, but generalizing this idea to phrases or sentences is less straightforward. 
In contrast, our proposed \methodname{} supports data augmentation in multiple granularity, and is able to reserve sequence order since drop operation does not change the relative order of the original input, which is important in many kinds of sequence data such as natural language.

%% file: appendix.tex
\subsection{Statistics of Benchmark Data Sets}

In this section, we summarize the statistics of the four natural language processing datasets used in our experiments in Table \ref{table:datastats}.
For SQuAD, since the dataset does not annotate supporting facts, we approximately estimate supporting facts by counting tokens that are part of a bigram that appears in the question.

\begin{table*}[!h]
    \centering
    \small
    \begin{threeparttable}
    \begin{tabular}{l|c |c |c |c}
        \toprule
        Data & Train & Dev & \# of Spans  & \# of Supporting facts  \\
        \midrule
        HotpotQA\citep{yang2018hotpotqa}\tnote{$\dagger$} & 90,564 &  \phantom{0}7,405 &  14.45/11/17/2/96\tnote{$\diamond$} & 2.38/2/3/2/12 \\
        MultiRC\citep{MultiRC2018}\tnote{$\dagger$} & \phantom{0}5,131 & \phantom{00,}953 & 14.72/12/18/6/41 & 2.32/2/2/2/6 \\
        DocRED\citep{yao2019docred}\tnote{$\dagger$} & 38,180 & 12,323 &8.24/6/10/3/25 & 1.67/1/2/1/11 \\
        SQuAD\citep{rajpurkar2016squad}\tnote{$\star$} & 87,599 & 10,570 & 156.26/114/186/25/853 & 7.74/3/11/0/82\tnote{$\ddagger$} \\
        \bottomrule
    \end{tabular}
    \label{tab:my_label}
    \begin{tablenotes}
    		\item [$\dagger$] Sentence-level span and supporting facts, and MultiRC is the second version and available at \url{https://cogcomp.seas.upenn.edu/multirc/}, where the column ``train'' and ``dev'' w.r.t. MultiRC reports the number of questions in training data and dev data, respectively.
			\item [$\star$] Token-level span, and the statistics of \# of spans/supporting facts are collected by setting span = 1 token;
			\item [$\diamond$] This is collected based on the top 4 documents selected from 10 candidate documents by document retriever. The five numbers correspond to Average, 25\% Percentile, 75\% Percentile, Min and Max, respectively.
			\item [$\ddagger$] We define the supporting facts as the spans of the context which have bigrams appearing in the question.
    \end{tablenotes}
    \end{threeparttable}
    \caption{Statistics of Benchmark Data sets}\label{table:datastats}
\end{table*}

\subsection{Experimental Setup for Different Tasks}
This section introduces the detailed implementations of our methods on four benchmark data sets, as well as the hyper-parameter setting for model optimization and baselines to compare with our implementations. 

\subsubsection{HotpotQA}
\xhdr{Implementation details} The objective of HotpotQA is answering questions from a set of 10 paragraphs where two paragraphs are relevant to the question and the rest are distractors. HotpotQA presents two tasks: answer span prediction and evidence sentence (i.e., supporting fact) prediction. Our HotpotQA model consists of two stages: the first stage selects top 4 paragraphs from 10 candidates by a retrieval model. The second stage finds the final answer span and evidence over the selected 4 paragraphs. We particularly feed the following input format to encoder: ``[CLS] question [SEP] sent$_{1,1}$ [SEP] sent$_{1,2}$ $\cdots$ [SEP] sent$_{4,1}$ [SEP] sent$_{4,2}$ $\cdots$ [SEP]''. And we apply the proposed span drop methods over all the sentences except for supporting facts.

For answer span prediction, we use the answer span prediction model in \citep{devlin2019bert} with an additional task of question type (yes/no/span) classification head over the first special token ([CLS]). For evidence extraction, we apply two-layer MLPs on top of the representations corresponding to sentence and paragraph to get the corresponding evidence prediction scores and use binary cross entropy loss to train the model. Finally, we combine answer span, question type, sentence evidence, and relevant paragraph losses and train the model in a multitask way using linear combination of losses. The hyper-parameter search space for our models on HotpotQA is given in Table~\ref{tab:hyper_setting}.

\xhdr{Baselines} We compare our implementation of HotpotQA model over Electra with the following strong baselines: 1) RoBERTa~\citep{liu2019roberta} based model, 2) long sequence encoder Longformer~\citep{beltagy2020longformer} based model and 3) SAE~\citep{tu2020select} which combines graph neural network and pretrained language models for multi-hop question answer and is the current SOTA model on HotpotQA. The numbers reported in our paper for the first two models come from \citep{zaheer2020big}.

\begin{table*}[!h]
    \centering
    \scriptsize
    \begin{tabular}{l|c |c|c |c}
    \toprule
      Parameter name & HotpotQA & MultiRC & DocRED & SQuAD \\
      \midrule
      Batch size   & \{4, 8\} & \{8, 16\} & \{4, 8\} & \{16, 32\} \\
      Learning rate   &\{3e-5, 2e-5, 1e-5, 1e-4\} & \{3e-5, 2e-5, 1e-5, 1e-4\} &  \{1e-4, 5e-5\} & \{1e-4, 5e-5\} \\
      Span Drop ratio & \{0.1, 0.15, 0.2, 0.25\} & \{0.1, 0.15, 0.2, 0.25\} & \{0.05, 0.1, 0.15, 0.2\} & \{0.03, 0.05, 0.1, 0.15, 0.2, 0.25\} \\
      Optimizer & AdamW & AdamW & AdamW & AdamW \\
      Epochs & 10 & 15 & \{10,20,30\} & \{2,4,6,8\} \\
      \bottomrule
    \end{tabular}
    \caption{Hyper-parameter search space for models on different benchmarks}
    \label{tab:hyper_setting}
\end{table*}

\subsubsection{MultiRC}

\xhdr{Implementation details} MultiRC is a multi-choice question answer task, which supplies a set of alternatives or possible answers to the question and requires to select the best answer(s) based on multiple sentences while a few of these sentences (i.e., supporting facts) are relevant to the questions. For given example of $k$ candidate answer and $n$ sentences, we first feed the following input to the encoder: ``[CLS] question [SEP] answer$_1$ [SEP] answer$_2$ [SEP] $\cdots$ answer$_k$ [SEP] sentence$_1$ [SEP] sentence$_2$ [SEP] $\cdots$ sentence$_n$ [SEP]''. For answer prediction, we apply two-layer MLPs on top of the representations corresponding to candidate answers and sentences to get the corresponding answer and sentences scores, and use a combination of two binary cross entropy losses to train the model in a multi-task way. We apply our span drop methods over all input sentences except for the evidence sentences. The hyper-parameter search space for our MultiRC models is given in Table~\ref{tab:hyper_setting}.  

\xhdr{Baselines} We compare our implementation of MultiRC model over Electra with the following baselines: 1) BERT based model, 2) RoBERTa~\citep{liu2019roberta} based model and 
3) REPT (RoBERTa-base) trained with new additional training tasks~\citep{jiao2021rept}.

\subsubsection{Relation extraction task: DocRED}
\xhdr{Implementation details} DocRED is document-level relation extraction which consists of two tasks: relation prediction of a given pair of entities and the evidence prediction. We construct the entity-guided inputs to the encoder following prior work \cite{huang2020entity}. Each training example is organized by concatenating the head entity, together with the $n$ sentences in the document as ``[CLS] head entity [SEP] sentence$_1$ [SEP] sentence$_2$ [SEP] $\cdots$ sentence$_n$ [SEP]''. 

For both relation extraction and evidence prediction, we apply a biaffine transformation that combines entity representations and entity/sentence representations, respectively, and score them using the adaptive threshold loss proposed by \citet{zhou2021atlop}.
We train the model in a multi-task setting by using a linear combination of relation extraction and evidence prediction losses. And we apply the proposed span drop methods over all input sentences except for those that serve as evidence for entity relations with the head entity. Please refer to Table~\ref{tab:hyper_setting} for the hyper-parameter search space of our DocRED models. 

\xhdr{Baselines} We compare against a set of strong baselines w.r.t. document-level relation including: 1) E2GRE~\citep{huang2020entity}, 2) ATLOP~\citep{zhou2021atlop} and 3) SSAN~\citep{xu2021entity}. 

\subsubsection{SQuAD}
\xhdr{Implementation details} SQuAD aims at extracting a segment of text from a given paragraph as answer to the question. We format the input example as ``[CLS] question [SEP] paragraph [SEP]'' and feed to the encoder. Following prior work, we employ a span prediction mechanism where the end of the span is conditioned on the representation of the start of the span, originally presented in the XLNet paper \citep{yang2019xlnet}.  And we apply our proposed span drop methods over the paragraph before feeding it into the model for training, where spans that contain question bigrams are preserved. The hyper-parameter space to optimize the SQuAD model is shown in Table~\ref{tab:hyper_setting}.

\xhdr{Baselines} We implemented models for SQuAD based on ELECTRA~\citep{clark2019electra} and compare them with existing model implementations for SQuAD based on 1) ELECTRA~\citep{clark2019electra}, 2) RoBERTa~\citep{liu2019roberta} and 3) XLNet~\citep{yang2019xlnet}, respectively.